\documentclass{article}
\usepackage{style}

\usepackage{times}
\usepackage{soul}
\usepackage{url}
\usepackage[hidelinks]{hyperref}
\usepackage[utf8]{inputenc}
\usepackage[small]{caption}
\usepackage{graphicx}
\usepackage{amsmath}
\usepackage{amsthm}
\usepackage{booktabs}
\usepackage{algorithm}
\usepackage{algorithmic}
\usepackage[switch]{lineno}
\usepackage{multirow} 
\usepackage{amssymb,amsmath,amsfonts}
\usepackage{pifont}
\newcommand{\xmark}{\ding{55}}
\usepackage[table]{xcolor}

\title{TSM-Pose: Topology-Aware Learning with Semantic Mamba for Category-Level Object Pose Estimation}

\author{
Jinshuo Liu$^1$
\and
Bingtao Ma$^1$\corresponding
\and
Junlin Su$^1$
\and
Guanyuan Pan$^1$
\and
Beining Wu$^1$
\and
\\Cheng Yang$^1$
\and
Jiaxuan Lu$^2$
\and
Chenggang Yan$^1$\correspondingmark
\And
Shuai Wang$^1$\correspondingmark
\affiliations
$^1$Hangzhou Dianzi University\\
$^2$Shanghai Artificial Intelligence Laboratory
}

\newcounter{corrauth}
\newcommand{\corresponding}{%
  \thanks{Corresponding Authors}%
  \setcounter{corrauth}{\value{footnote}}%
}
\newcommand{\correspondingmark}{\footnotemark[\value{corrauth}]}

\begin{document}

\maketitle

\begin{abstract}
Category-level object pose estimation is fundamental for embodied intelligence, yet achieving robust generalization to unseen instances remains challenging. However, existing methods mainly rely on simple feature extraction and aggregation, which struggle to capture category-shared topological structures and conduct semantic keypoint modeling, limiting their generalization. To address these, we propose a \textbf{T}opology-Aware Learning with \textbf{S}emantic \textbf{M}amba for Category-Level \textbf{P}ose Estimation framework (TSM-Pose). Specifically, we introduce a Topology Extractor to capture the global topological representation of the point cloud, which is integrated into local geometry features and enables robust category-level structural representation. Simultaneously, we propose a Mamba-based Global Semantic Aggregator that injects semantics priors into keypoints to enhance their expressiveness and leverages multiple TwinMamba blocks to model long-range dependencies for more effective global feature aggregation. Extensive experiments on three benchmark datasets (REAL275, CAMERA25, and HouseCat6D) demonstrate that TSM-Pose outperforms existing state-of-the-art methods.
\end{abstract}

\section{Introduction}
Object pose estimation stands as a cornerstone of Embodied AI and ubiquitous robotics, enabling machines to perceive and interact with the physical world through tasks such as robotic grasping and manipulation~\cite{sun2025review,wen2022you,wu2020grasp,cong2021comprehensive}, autonomous driving~\cite{hoque2023deep,mao20233d,sun2021data}, and 3D scene understanding~\cite{zhu2025move,corsetti2025functionality,ma2023topology,10215502}. Traditional instance-level paradigms~\cite{xiang2017posecnn} often struggle due to their reliance on high-fidelity CAD models, which limit their utility in the open world, where encountering novel, unmodeled objects is the norm rather than the exception.
To address this issue, researchers proposed a \textbf{C}ategory-level \textbf{O}bject \textbf{P}ose \textbf{E}stimation~\cite{wang2019normalized} (COPE), which can estimate the pose and dimensions of unseen instances without relying on CAD models. These methods typically offer greater flexibility and stronger generalization ability. However, COPE remains challenging due to significant intra-class shape variations and complex real-world observation conditions (\emph{e.g.}, occlusion and noise).

Most COPE methods~\cite{yu2024catformer,chen2022stereopose,tian2020shape} employ the dense correspondence paradigm, which establishes point-to-point correspondences and recovers poses by learning to map RGB-D observations to a normalized object coordinate space (NOCS)~\cite{wang2019normalized}. However, when there are significant geometric differences within a category, the lack of stable one-to-one correspondences between instances reduces generalization ability and prediction stability for unseen instances. Recent research~\cite{yu2025keypose,xu2025pre,lin2024instance} has gradually shifted towards sparse-keypoint-based strategies. By abstracting objects into semantically consistent keypoints, these methods aim to construct a universal structural skeleton that transcends individual geometry, enhancing robustness and generalization under shape variations and complex observation conditions.

\begin{figure}[t]
\centering
\includegraphics[width=\linewidth]{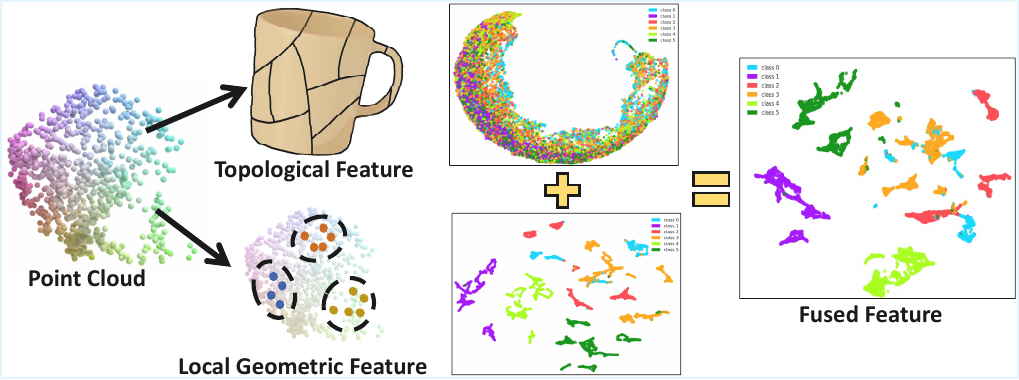}
\caption{
TUMAP visualizations of local geometric features (PointNet++), topological features, and their fusion.
The proposed Topology Extractor derives topology-aware representations from point clouds. We integrate global topological features into local geometric features, leading to more compact intra-class feature distributions (\emph{i.e.}, fused features) while preserving inter-class separability.
}
\label{feature}
\end{figure}

AG-Pose~\cite{lin2024instance} is a typical keypoint-based method that improves cross-instance keypoint stability via instance-adaptive detection and geometry-aware feature aggregation. However, its feature extraction and aggregation capabilities are still limited when faced with significant intra-class deformations: the features used are mainly local features, which are easily affected by deformations, and its simple pooling operation lacks the ability to model the global structure, making it difficult to capture long-range dependencies between key points. Meanwhile, compressing dense point clouds into sparse keypoints inevitably causes semantic dilution, leading to the loss of category-level context required for global pose reasoning and thereby reducing its robustness.

To address these challenges, we propose TSM-Pose, a topology-aware, semantic-enhanced framework for COPE that explicitly integrates structural and semantic priors into keypoint-based pose estimation. To overcome the inherent sensitivity of local geometric features to intra-class deformations, we introduce a Topology Extractor that derives topology-aware representations from point clouds, complementing local geometric features for more stable structural modeling, as shown in Figure~\ref{feature}.
To aggregate sparse keypoints with long-range structural reasoning, we design an Mamba-based Global Semantic Aggregator (MGSA) that leverages TwinMamba to model global dependencies among keypoints and enhances holistic feature aggregation.
Moreover, to mitigate semantic context loss introduced by keypoint compression, we innovate a Semantic Injection Mechanism, which appends a lightweight category-encoded CAT token to the keypoint sequence, injecting explicit semantic priors and allowing category-level information to be progressively propagated during global sequence modeling. Extensive experiments on CAMERA25~\cite{wang2019normalized}, REAL275~\cite{wang2019normalized}, and HouseCat6D~\cite{jung2024housecat6d} demonstrate that TSM-Pose  achieves state-of-the-art performance.

We outline the contributions of TSM-Pose as follows:

\begin{itemize}
    \item We propose TSM-Pose, a novel COPE framework that enhances the robustness and generalization of pose estimation in scenarios with significant intra-class shape differences by improving global structural representation and keypoint sequence semantic modeling capabilities. It achieves SOTA performance on the benchmark datasets CAMERA25, REAL275, and HouseCat6D.
    \item We introduce topology Exactor to obtain deformation-insensitive global structural representations from point clouds, effectively compensating for the shortcomings of traditional local geometric features in cross-instance global structural modeling.
    \item We design an MGSA to model long-range geometric dependencies among keypoints while preserving structural consistency. In addition, we introduce category Semantic Injection to propagate category priors and alleviate semantic loss caused by sparse keypoints.

\end{itemize}

\begin{figure*}[t]
\centering
\includegraphics[width=\linewidth]{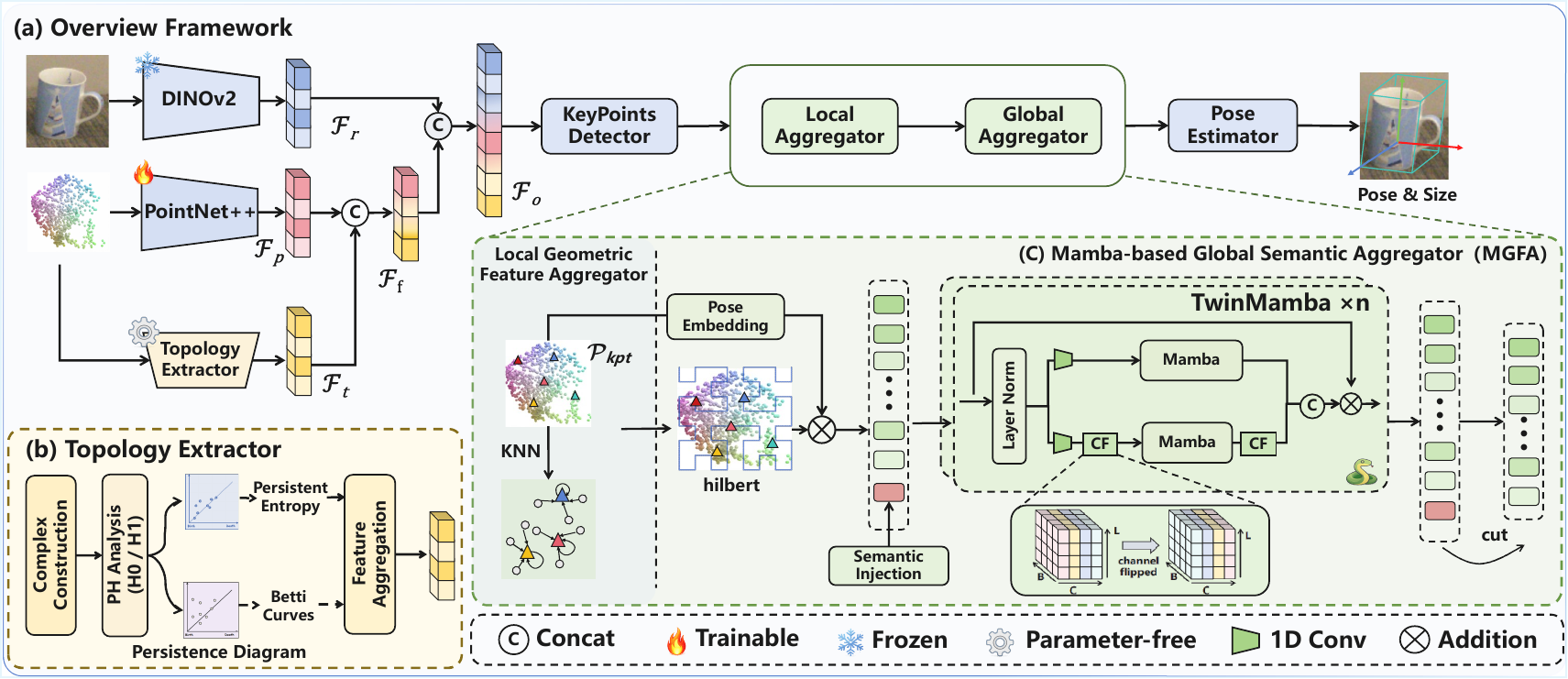}
\caption{
(a) Overall framework of TSM-Pose. We fuse multimodal features \(\mathcal{F}_r,\mathcal{F}_p\) and \(\mathcal{F}_t\) into \(\mathcal{F}_o\), which are then fed into a keypoint detector. The detected keypoints, along with their associated features, are subsequently
passed to a Local geometric feature Aggregator (LGFA) and MGSA. Finally, we use a pose estimator to predict the object's pose. 
(b) Topology Extractor capturing intrinsic topological structures of the point cloud using persistent homology with homology dimensions \(H_0\) and \(H_1\).
(c) MGSA serializes the key points and performs Semantic Injection, then models long-range dependencies through stacked TwinMamba blocks.
}
\label{overview}
\end{figure*}

\section{Related Work}
\subsection{Category-level Object Pose Estimation}

COPE aims to predict the 6D pose of unseen instances within the same object category without relying on instance-level CAD models. We categorize existing approaches into shape-prior-based methods and shape-prior-free methods. Early studies primarily focused on explicit shape prior modeling. NOCS~\cite{wang2019normalized} introduced a unified normalized coordinate space that maps category instances into a canonical reference frame, enabling joint estimation of object pose and size by the Umeyama algorithm. Subsequent works~\cite{Zheng_GeoReF_2024_CVPR,liu2024mh6d,li2025gce} further built upon category templates or canonical shapes by incorporating deformation modeling, geometric alignment, or correspondence learning to handle complex intra-class geometric variations. In recent years, researchers have proposed a series of shape-prior-free methods. SpotPose~\cite{ren2025rethinking} improves performance by using explicit anomaly key identification and clearing. AG-Pose~\cite{lin2024instance} introduces adaptive geometry-aware keypoint learning. MK-Pose~\cite{yang2025mk} explores Self-supervised keypoint learning with polymorphic fusion. TSM-Pose also follows the shape-prior-free paradigm. However, there are still performance bottlenecks in terms of estimation accuracy and robustness that need further improvement.

\subsection{Mamba in Point Cloud Tasks}
Mamba is an efficient sequence modeling architecture that excels at modeling long-range dependencies through selective state updates, while maintaining high performance and throughput, and has recently attracted significant research attention. Since the original Mamba follows a causal modeling paradigm, subsequent studies (e.g., Vision Mamba) have further explored bidirectional Mamba to enhance information interaction at both ends of the sequence, thereby improving global modeling capability~\cite{vim}. Motivated by these advances, recent studies have begun exploring the application of Mamba to 3D point cloud understanding tasks. PointMamba~\cite{liang2024pointmamba} maps unordered point clouds into 1D sequences using the Hilbert space-filling curve and its Trans-Hilbert variant, and applies Mamba to perform global feature modeling while preserving spatial continuity as much as possible. Furthermore, Mamba3D~\cite{han2024mamba3d} highlights the importance of local geometric priors in point clouds and introduces a bi-SSM design to improve point cloud representation. More recently, INKL-Pose~\cite{zhang2025instance} incorporates Mamba-based global modeling into category-level object pose estimation and demonstrates its effectiveness. These studies suggest that Mamba has strong potential for efficient long-range modeling in 3D point cloud understanding and pose estimation. Inspired by these advances, we leverage the global modeling capability and favorable efficiency–performance trade-off of Mamba to further enhance COPE.

\subsection{Topology-Aware Point Cloud Analysis}

There has been a growing interest in incorporating topological information~\cite{su2025topological} into point cloud processing. These approaches aim to extract global structural properties inherent in point clouds, such as connectivity, holes, cavities, and branching structures, which can be characterized as topological invariants. Researchers employ tools from topological data analysis (TDA)~\cite{carlsson2009topology}, such as persistent homology, to model structural features that remain stable across multiple scales. 
Such global structural information is often difficult for neural networks that rely on local receptive fields to learn effectively. Moreover, topological features exhibit strong robustness to point cloud noise and non-uniform sampling. While existing research explores the application of topological descriptors for point cloud reconstruction~\cite{jignasu2024sdfconnect} and classification \cite{ghosh2025taco,chen2025robust}, the effective integration of topological invariants with learnable geometric features within the context of COPE remains largely unexplored. Therefore, incorporating topological features into point cloud representation learning offers a promising approach to enhancing COPE's cross-instance generalization ability.

\section{Method}
\subsection{Overview}
Following previous works~\cite{lin2024instance,yang2025mk}, we apply Mask R-CNN for instance segmentation to obtain cropped RGB image \(\mathcal{I}_{o}\in\mathbb{R}^{H\times W\times 3}\) and object point cloud \(\mathcal{P}_{o}\in\mathbb{R}^{N\times 3}\), where N is the number of points,  as shown in Figure~\ref{overview}. TSM-Pose  uses them to estimate the 3D rotation \(\mathbf{R}\in\mathbb{R}^{3\times 3}\), translation \(\mathbf{t}\in\mathbb{R}^{3}\), and object size \(\mathbf{s}\in\mathbb{R}^{3}\). 

TSM-Pose consists of four stages. First, we extract visual, geometric, and topological features from the image and point cloud, and fuse them into a unified point-by-point feature representation to serve as input for subsequent keypoint inference. Second, we employ an  keypoint detector to predict a set of sparse yet discriminative keypoints. Subsequently, we feed the predicted keypoints and their corresponding features from the fused features into the LGFA and MGSA, thereby enhancing the keypoint representation by fusing local geometric details and global semantic context. Finally, the improved keypoint features are used to regress the object's pose.

\subsection{Feature Extractor}
We obtain visual representations from the RGB image \(\mathcal{I}_{o}\) by DINOv2 (ViT-S/14)~\cite{oquab2023dinov2}. Meanwhile, We use a PointNet++~\cite{qi2017pointnet++} to extract local geometric features \(\mathcal{F}_p \in \mathbb{R}^{N \times d_p}\) and a Topology Extractor to extract and obtain global topological features \(\mathcal{F}_{t}\in \mathbb{R}^{d_t}\). We then broadcast \(\mathcal{F}_{t}\) to the point level and fuse it with \(\mathcal{F}_{p}\) to obtain \(\mathcal{F}_{f} \in \mathbb{R}^{N \times d_f}\). We project each 3D point onto the image and fetch its pixel embedding by bilinear sampling to obtain
\(\mathcal{F}_I \in \mathbb{R}^{N \times d_r}\). Then we concatenate \(\mathcal{F}_I\) with \(\mathcal{F}_{f}\) to build
\(\mathcal{F}_{o} \in \mathbb{R}^{N \times d}\), which we use as the network input.

\subsection{Topology Extractor}

In most COPE frameworks~\cite{lin2024instance,cvpr2025spotpose}, learning-based geometric representations (such as PointNet++) can be sensitive to variations in point cloud sampling density, occlusion, and appearance differences between instances within the same class. This sensitivity can easily lead to the separation of feature distributions within instances of the same class (as shown in Figure~\ref{feature}). In contrast, topological structures exhibit greater stability to these perturbations and can provide global shape constraint information beyond local geometric details. Based on this, we introduce a TDA-inspired topological structure descriptor to supplement the shortcomings of local geometric features.

To capture the global topological structure, we construct an Alpha complex from \(\mathcal{P}_{o}\) using a scale parameter \(\alpha\). 
Specifically, for each point \(\mathcal{P}_{o}^i\), we define a closed ball \(B(\mathcal{P}_{o}^i, \alpha)\) centered at \(\mathcal{P}_{o}^i\) with radius \(\alpha\). 
We include a \(k\)-simplex formed by points \(\{\mathcal{P}_{o}^{i_0}, \ldots, \mathcal{P}_{o}^{i_N}\}\) in the Alpha complex~\cite{edelsbrunner1994three} at scale \(\alpha\) if the corresponding balls have a non-empty intersection:
\begin{equation}
\bigcap_{j} B(\mathcal{P}_{o}^{i_j}, \alpha) \neq \emptyset .
\end{equation}
As \(\alpha\) increases, we obtain a series of nested alpha complexes, forming an alpha-filtering process. Based on this filtering, we compute its
\(k\)-dimensional persistence~\cite{edelsbrunner2002topological}:
\begin{equation}
\mathcal{D}_k = \left\{ (b_j^{(k)}, d_j^{(k)}) \right\}, 
\qquad k \in \{0,1\}.
\end{equation}
Here, \(b_j^{(k)}\) and \(d_j^{(k)}\) are the birth and death scales of the j-th k-th topological feature during the filtering process. We use $j$ as indices to enumerate all topological features in the persistence diagram $\mathcal{D}_k$. We define the persistence length of the \(j\)-th feature as \(\ell_j^{(k)} = d_j^{(k)} - b_j^{(k)}\). Let \(p_j^{(k)} = \frac{\ell_j^{(k)}}{\sum_i \ell_i^{(k)}}\), then we denote the topological entropy as follows:
\begin{equation}
H^{(k)} = -\sum_{j} p_j^{(k)} \log\left(p_j^{(k)} + \varepsilon\right).
\end{equation}
Here, \(\varepsilon\) is a small constant for numerical stability.

The entropy characterizes the distribution of lifetimes of topological features in the persistence diagram, quantifying the relative importance and diversity of global topological structures within the shape. To better capture the evolution of topology across different scales, we linearly normalize all birth and death values into the range \([0,1]\), resulting in \((\tilde b_j^{(k)}, \tilde d_j^{(k)})\). We then uniformly sample scales \(t_m \in [0,1]\) and count the number of features that remain alive at each scale, which results in the Betti curve:
\begin{equation}
\beta^{(k)}(t_m) =
\#\left\{ j \;\middle|\; \tilde b_j^{(k)} \le t_m < \tilde d_j^{(k)} \right\},
\end{equation}
\(\#\{\cdot\}\) denotes the cardinality of a set.
We incorporate a global topological feature vector consisting of topological entropy and Betti curves into the point cloud representation to obtain topology-enhanced features \(\mathcal{F}_{t}\in \mathbb{R}^{N_k \times {d_t}}\).  
Then, keypoints $\mathcal{F}_{kpt}$ are detected by multimodal features \(\mathcal{F}_o\) and the
instance-adaptive keypoint detector~\cite{lin2024instance}. The further details are available in the supplementary material.

\subsection{Local Geometric Feature Aggregator}
Given the detected keypoints, we explicitly aggregate their neighborhood feature with a Local Geometric Feature Aggregator (LGFA), which based on graph attention mechanism~\cite{velickovic2017graph} to better capture local geometric structure.
Let $\mathbf{f}_m \in \mathcal{F}_{kpt}$ denote the feature of the $m$-th keypoint, and let $\mathcal{N}(m)$ represent its $K$ nearest neighbors (KNN).
We encode Local geometric information via the relative displacement 
$\Delta \mathbf{p}_{m,j} = \mathbf{p}_j - \mathbf{p}_m$ using a geometric encoder:
\begin{equation}
\mathbf{e}_{m,j} = MLP_{geo}(\Delta \mathbf{p}_{m,j}).
\end{equation}
To model the dynamic relationship among a keypoint, its neighbors, and their relative geometry, we compute attention coefficients from a concatenation of the query, key, and geometric representations. We obtain the attention score $c_{m,j}$~as
\begin{equation}
\mathbf{c}_{m,j} = \mathrm{LeakyReLU}\left( 
\mathbf{a}^T [ \mathbf{q}_m \,;\, \mathbf{k}_j \,;\, \mathbf{e}_{m,j} ] 
\right).
\end{equation}
We obtain the attention weights $\alpha_{m,j}$ via softmax normalization over the local neighborhood. Using this, we aggregate neighborhood features by incorporating the geometric encoding into the value representations:
\begin{equation}
\mathbf{f}_m^{\mathrm{loc}} = 
\sum_{j \in \mathcal{N}(m)} \alpha_{m,j} (\mathbf{v}_j + \mathbf{e}_{m,j}).
\end{equation}
We then fuse the aggregated local feature with the original keypoint feature via a residual connection:
\begin{equation}
\mathbf{\hat{f}}_{kpt} =
\mathrm{ReLU}(\mathbf{f}_{kpt}^{\mathrm{loc}} + \mathbf{f}_{kpt}).
\end{equation} Stacking all keypoints yields the output feature set \(\hat F_{kpt}\).

\subsection{Mamba-based Global Semantic Aggregator}

Although LGFA effectively captures fine-grained geometric information, it lacks the capacity to model the global structure required for COPE. Existing keypoint-based methods~\cite{lin2024instance,yang2025mk} use simple pooling and concatenation for global feature aggregation, which not only loses substantial local keypoint semantic context but also ignores the global geometric relationships between keypoints. To address these issues, we propose MGSA to enhance the global semantic representation of keypoints by introducing an Semantic Injection and combining TwinMamba to model long-range dependencies between keypoints.

\paragraph{Keypoint Serialization.}Unlike discriminative point cloud tasks that emphasize permutation invariance, keypoints in pose estimation are highly sparse, making ensemble modeling insufficient for capturing global geometry. 
We serialize keypoints using a single Hilbert space-filling curve, providing a stable traversal order while preserving local spatial continuity. We formulate the one-dimensional keypoint sequence \(\mathcal{H}_{kpt}=
\big[\mathbf{h}_1;\ \mathbf{h}_2;\ \ldots;\ \mathbf{h}_{N_k}\big]\in\mathbb{R}^{N_k\times d},\) where $\mathbf{h}_{N_k}$ denotes the feature embedding of the ${N_k}$-th keypoint after local feature aggregation.

\paragraph{Semantic Injection.}  To address the loss of inherent semantic context in point features, we inject category semantics into the keypoint sequence to augment local keypoints with semantic information, which enables category-level semantics to be progressively propagated along the keypoint sequence, allowing subsequent Mamba networks to perform global modeling conditioned on prior semantic information. Furthermore, we encode the category semantics into an independent global semantic token and place it at the beginning of the sequence to provide a stable semantic anchor: \(\mathbf{c}=\mathbf{MLP}(\mathrm{CAT}_n),\quad n\in[0,5], \tilde{\mathcal{H}}_{kpt}=\big[\mathbf{c};\mathbf{h}_1;\mathbf{h}_2;\ldots;\mathbf{h}_{N_k}\big]\in\mathbb{R}^{(N_k+1)\times d}.\) Subsequently, we input the sequence $\tilde{\mathcal{H}}_{kpt}$ into TwinMamba for global modeling.

\begin{figure}[t]
    \centering
    \setlength{\tabcolsep}{0.5mm} 
    \renewcommand{\arraystretch}{1} 

    \begin{tabular}{@{}c@{\hspace{3mm}}cccc@{}}
        \rotatebox{90}{\textbf{AG-Pose}} &
        \includegraphics[width=0.22\linewidth]{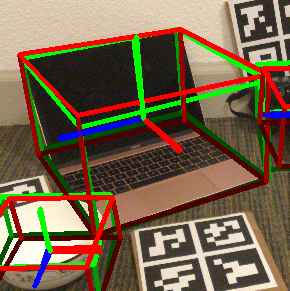} &
        \includegraphics[width=0.22\linewidth]{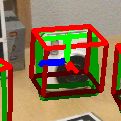} &
        \includegraphics[width=0.22\linewidth]{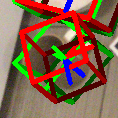} &
        \includegraphics[width=0.22\linewidth]{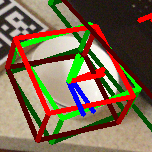} \\[0.5mm]

        \rotatebox{90}{\textbf{TSM-Pose}} &
        \includegraphics[width=0.22\linewidth]{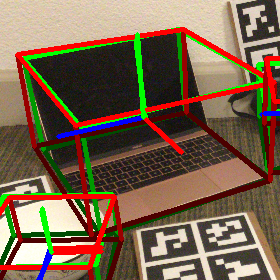} &
        \includegraphics[width=0.22\linewidth]{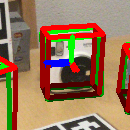} &
        \includegraphics[width=0.22\linewidth]{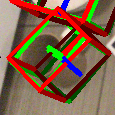} &
        \includegraphics[width=0.22\linewidth]{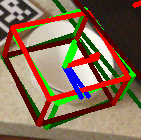} \\
    \end{tabular}

\caption{We visualized and compared TSM-Pose and our baseline AG-Pose on the REAL275 dataset, where predicted and ground truth bounding boxes are represented in green and red, respectively.}
\label{visual_comparision}
\end{figure}

\paragraph{TwinMamba Block.}
Since sparse keypoints constitute a high-level abstraction of object geometry, effectively capturing their intricate spatial dependencies is crucial. To fully leverage this relational modeling potential, following previous Mamba-based works~\cite{han2024mamba3d,zhang2025instance}, we employ a TwinMamba module to implement global keypoint modeling.

To alleviate the model's over-reliance on serialization order, we construct two complementary sequence branches. Specifically, one branch models keypoints along the original Hilbert curve sequence, while the other branch first applies a channel flipping (CF) operation while preserving the keypoint order, and then feeds the flipped features into a complete Mamba block for modeling. In this way, a symmetric modeling path is constructed along the feature dimension.
Given a keypoint feature vector:
\[
\mathbf{\hat{f}}_{N_k} =
\big[ f_{N_k,1},\, f_{N_k,2},\, \ldots,\, f_{N_k,d} \big] \in \mathbb{R}^{d},
\]
CF performs a channel-wise reversal:
\[
CF(\mathbf{\hat{f}}_{N_k}) =
\big[ f_{N_k,d},\, f_{N_k,d-1},\, \ldots,\, f_{N_k,1} \big] \in \mathbb{R}^{d}.
\]
We can uniformly express the two branches as:
\begin{equation}
\hat{\mathcal{H}}_f=\mathrm{Mamba}(\tilde{\mathcal{H}}_{kpt}), \quad
\hat{\mathcal{H}}_b=\mathrm{CF}\big(\mathrm{Mamba}(\mathrm{CF}(\tilde{\mathcal{H}}_{kpt}))\big),
\end{equation} 
Subsequently, we input the two sequences into a parameter-shared Mamba module for global modeling. Finally, we fuse the two branch outputs at the token level, followed by feature refinement through residual connections:
\begin{equation}
\hat{\mathbf{h}}_k=
\mathbf{h}_k+
\mathrm{MLP}\!\left(
\left[\hat{\mathbf{h}}_{f,k}\ \Vert\ \hat{\mathbf{h}}_{b,k}\right]
\right).
\end{equation}
 We then feed the updated keypoint representation \(\hat{\mathcal{H}}_{kpt} \in \mathbb{R}^{N_k \times d}\) into the subsequent pose regression module.

To ensure that the aggregated keypoints retain meaningful geometric structures, we employ a reconstruction-based supervision mechanism~\cite{lin2024instance}. We reconstruct the shape by predicting multiple 3D offset vectors for each keypoint and adding them to the corresponding keypoint coordinates, yielding a set of local point clusters whose union forms the reconstructed point cloud. 
The reconstructed point cloud is denoted as $\mathcal{P}^{recon}_{o}$ and the ground-truth point cloud is $\mathcal{P}^{gt}_{o}$. 
We supervise the reconstruction using a symmetric Chamfer Distance (CD) to enforce geometric consistency:
\begin{equation}
\mathcal{L}_{recon} = \mathrm{CD}(\mathcal{P}^{recon}_{o}, \mathcal{P}^{gt}_{o}).
\end{equation}
To prevent excessive deformation, we further regularize the predicted offset vectors $\{\Delta_i\}$ with an $\ell_2$ penalty:
\begin{equation}
\mathcal{L}_{\delta} = \frac{1}{N}\sum_{i=1}^{N}\|\Delta_i\|_2.
\end{equation}

\begin{table*}[t]
\centering
\caption{Quantitative comparison on the REAL275 dataset against SOTA approaches. '*' denotes CATRE\protect\cite{liu_2022_catre} IoU metrics, and '–' denotes unavailable entries. We emphasize the top and runner-up results in \textbf{bold} and \underline{underlined}.}
\label{comparison}

\renewcommand{\arraystretch}{1.15}
\resizebox{\textwidth}{!}{
\begin{tabular}{l c c rrrrr}
\toprule
\textbf{Methods} & \textbf{Venue/Source} & \textbf{Shape Prior}
& \(\mathbf{IoU^{*}_{75}}\)
& \(\mathbf{5^\circ\,2\,\mathrm{cm}}\)
& \(\mathbf{5^\circ\,5\,\mathrm{cm}}\)
& \(\mathbf{10^\circ\,2\,\mathrm{cm}}\)
& \(\mathbf{10^\circ\,5\,\mathrm{cm}}\) \\
\midrule
MH6D~\cite{liu2024mh6d} & TNNLS'24 & \checkmark & 54.2 & 53.0 & 61.1 & 72.0 & 82.0 \\
GeoRef~\cite{Zheng_GeoReF_2024_CVPR} & TNNLS'24 & \checkmark & 51.8 & 54.4 & 60.3 & 71.9 & 79.4 \\
CatFormer~\cite{yu2024catformer} & AAAI'24 & \xmark & -- & 47.7 & 53.7 & 69.0 & 79.5 \\
CLIPose~\cite{lin2024clipose} & TCSVT'24 & \xmark & -- & 48.5 & 58.2 & 70.3 & 85.1 \\
SpotPose~\cite{ren2025rethinking} & CVPR'24 & \xmark & -- & 59.7 & 68.3 & 76.5 & 84.2 \\
AG-Pose~\cite{lin2024instance} & CVPR'24 & \xmark & 61.3 & 57.0 & 64.6 & 75.1 & 84.7 \\
GCE-Pose~\cite{li2025gce} & CVPR'25 & \checkmark & -- & 57.0 & 65.1 & 75.6 & 86.3 \\
KeyPose~\cite{yu2025keypose}  & AAAI'25 & \xmark & -- & 57.7 & 66.0 & 78.8 & 88.0 \\
SpherePose~\cite{ren2025learning} & ICLR'25 & \xmark & -- & 58.2 & 67.4 & 76.2 & \underline{88.2} \\
MK-Pose~\cite{yang2025mk} & IROS'25 & \xmark & -- & 60.8 & -- & 78.0 & 84.6 \\
CleanPose~\cite{lin2025cleanpose} & ICCV'25 & \xmark & \underline{62.7} & \underline{61.7} & \underline{67.6} & \underline{78.3} & 86.3 \\
\midrule
\textbf{TSM-Pose (ours)} & -- & \xmark & \textbf{63.1} & \textbf{62.4} & \textbf{68.5} & \textbf{81.5} & \textbf{88.7} \\
\bottomrule
\end{tabular}}
\end{table*}

\begin{table}[t]
\centering
\caption{Quantitative comparison on the CAMERA25 dataset against SOTA approaches. '*' denotes CATRE\protect\cite{liu_2022_catre} IoU metrics, and '–' denotes unavailable entries. The top and runner-up results are emphasized in \textbf{bold} and \underline{underlined}.}
\label{camera_comparison}
\renewcommand{\arraystretch}{1.2}
\resizebox{\columnwidth}{!}{
\begin{tabular}{l rrrrr}
\toprule

\textbf{Methods} 
& \(\mathbf{IoU^{*}_{75}}\)
& \(\mathbf{5^\circ\,2\,\mathrm{cm}}\)
& \(\mathbf{5^\circ\,5\,\mathrm{cm}}\)
& \(\mathbf{10^\circ\,2\,\mathrm{cm}}\)
& \(\mathbf{10^\circ\,5\,\mathrm{cm}}\)
\\
\midrule
MH6D~\cite{liu2024mh6d}   & 54.2 & 53.0 & 61.1 & 72.0 & 82.0 \\
GeoRef~\cite{Zheng_GeoReF_2024_CVPR}   & 79.2 & 77.9 & 84.0 & 83.8 & 90.5 \\
VI-Net~\cite{lin2023vi}   & - & 74.1 & 81.4 & 79.3 & 87.3 \\
CLIPose~\cite{lin2024clipose}   & - & 74.8 & 82.2 & 82.0 & 91.2 \\
SpotPose~\cite{ren2025rethinking}   & - & 80.4 & 83.8 & 87.7 & 92.2 \\
AG-Pose~\cite{lin2024instance}   & \textbf{81.2} & 79.5 & 83.7 & 87.1 & 92.6 \\
MK-Pose~\cite{yang2025mk}  & - & 77.9 & - & 86.1 & 91.7 \\
KeyPose~\cite{yu2025keypose}   & - & 79.8 & 83.6 & 87.1 & 92.3  \\
CleanPose~\cite{lin2025cleanpose}    & 80.7 & \textbf{80.3} & \underline{84.2} & \underline{87.7} & \underline{92.7} \\
\midrule
\textbf{TSM-Pose (ours)}   & \underline{81.1} & 
\underline{79.9} & \textbf{84.6} & \textbf{88.2} & \textbf{93.6}\\
\bottomrule
\end{tabular}
}

\end{table}

\begin{table}[t]
\centering
\caption{Quantitative comparison on the HouseCat6D dataset against SOTA approaches. The top and runner-up results are emphasized in \textbf{bold} and \underline{underlined}.}
\label{HouseCat6D_comparison}
\renewcommand{\arraystretch}{1.3}

\resizebox{\columnwidth}{!}{
\begin{tabular}{l rrrrr}
\toprule

\textbf{Methods} 
& \(\mathbf{IoU_{25}}\)
& \(\mathbf{5^\circ\,2\,\mathrm{cm}}\)
& \(\mathbf{5^\circ\,5\,\mathrm{cm}}\)
& \(\mathbf{10^\circ\,2\,\mathrm{cm}}\)
& \(\mathbf{10^\circ\,5\,\mathrm{cm}}\)
\\
\midrule

VI-Net~\cite{lin2023vi}   & 80.7 & 8.4 & 10.3 & 20.5 & 29.1 \\
SecondPose~\cite{chen2024secondpose}   & 83.7 & 11.0 & 13.4 & 25.3 & 35.7 \\
AG-Pose~\cite{lin2024instance}   & 88.1 & 21.3  &22.1 & 51.3  & 54.3 \\

SpherePose~\cite{ren2025learning} & 88.8 & 19.3 & \textbf{25.9}& 40.9 & 55.3 \\

CleanPose~\cite{lin2025cleanpose} & \underline{89.2} & \underline{22.4} & 24.1 & \underline{51.6} & \underline{56.5} \\

\midrule
\textbf{TSM-Pose (ours)}   & \textbf{90.2} & 
\textbf{23.6} & \underline{25.1} & \textbf{53.2} & \textbf{57.8}\\
\bottomrule
\end{tabular}
}

\end{table}

\subsection{Loss Function}
We follow prior work~\cite{lin2024instance} and jointly supervise pose estimation, keypoint localization, NOCS prediction, and shape reconstruction with a multi-term loss:
\begin{equation}
\begin{aligned}
\mathcal{L}
= {} & \lambda_{pose}\mathcal{L}_{pose}
+ \lambda_{nocs}\mathcal{L}_{nocs}
+ \lambda_{cd}\mathcal{L}_{cd} \\
& + \lambda_{div}\mathcal{L}_{div}
+ \lambda_{recon}\mathcal{L}_{recon}
+ \lambda_{\delta}\mathcal{L}_{\delta},
\end{aligned}
\end{equation}
where $\mathcal{L}_{pose}$ is an \(\ell_2\) regression loss for predciting rotation \(\mathbf{R} \in \mathbb{R}^{3 \times 3}\), translation \(\mathbf{t} \in \mathbb{R}^{3}\), and scale \(\mathbf{s} \in \mathbb{R}^3\) of aligning an object instance to the NOCS. $\mathcal{L}_{pose}$ is formulated as:
\begin{equation}
\mathcal{L}_{pose}
= \|R - R_{gt}\|_2
+ \|t - t_{gt}\|_2
+ \|s - s_{gt}\|_2,
\end{equation}
where \(\mathbf{R}_{gt}\), \(\mathbf{t}_{gt}\), and \(\mathbf{s}_{gt}\) denote the ground-truth rotation, translation, and scale, respectively. 
In addition, \(\mathcal{L}_{nocs}\) enforces consistency between predicted keypoint NOCS coordinates and the ground truth.
\(\mathcal{L}_{cd}\) encourages keypoints to lie on the object surface, while \(\mathcal{L}_{div}\) promotes their spatial diversity.
Definitions and implementation details of the remaining loss terms are given in the supplementary materials.

\section{Experiment}

\subsection{Dataset}

In the experiment, we conduct evaluations on two widely used datasets, NOCS~\cite{wang2019normalized} and HouseCat6D~\cite{jung2024housecat6d}. NOCS contains a synthetic split (CAMERA25) and a real-world split (REAL275). CAMERA25 includes 275K synthetic RGB-D images for training and 25K for testing, while REAL275 includes 4,318 real images for training and 2,754 for testing. It covers six common categories (bottle, bowl, camera, can, laptop, and mug) with accurate pose and size annotations. HouseCat6D~\cite{jung2024housecat6d} is a large-scale multimodal pose dataset with 10 household-object categories and 194 object instances, featuring diverse viewpoints, heavy occlusion, and realistic scenes.

\subsection{Evaluation Metrics}
Following prior works~\cite{lin2024instance,lin2025cleanpose}, we evaluate the performance using two metrics. \textbf{3D IoU:} This metric measures the spatial overlap between the predicted and ground-truth 3D bounding boxes.
We adopt the 3D IoU defined in CATRE~\cite{liu_2022_catre} and report the mean Average Precision under a threshold of 75\%.
For the HouseCat6D~\cite{jung2024housecat6d} dataset, we additionally report the mAP of 3D IoU under thresholds of 25\% and 50\%.
\textbf{n°m cm:} This metric jointly evaluates rotation and translation errors for pose estimation.
A prediction counts as correct if the rotation error is below \(n^\circ\) and the translation error is below \(m\) cm.
Following standard evaluation protocols, we report results under four commonly used thresholds across all datasets:
\(5^\circ\)2 cm, \(5^\circ\)5 cm, \(10^\circ\)2 cm, and \(10^\circ\)5 cm.

\subsection{Implementation Details}
In the experiment, we uniformly sample each point cloud to \(N=1024\) points and normalize them.
During training, we add Gaussian noise to the point clouds and apply color jittering and random noise perturbations to the RGB images.
We set the input RGB image resolution to \(224 \times 224\) and the number of keypoints to \(N_k=96\).
We construct local neighborhoods using the KNN with \(K=16\).
The feature dimensions are \(d_t=96\), \(d_r=d_p=128\), and \(d=256\).
We extract Topological features corresponding to the \(0\)- and \(1\)-dimensional Betti numbers. The loss function is a weighted sum of multiple terms with
\(\lambda_{\text{pose}}=0.3\),
\(\lambda_{\text{nocs}}=2.0\),
\(\lambda_{\text{cd}}=2.0\),
\(\lambda_{\text{div}}=10\),
\(\lambda_{\text{recon}}=15\),
\(\lambda_{\delta}=1.0\). We conduct Training on a single NVIDIA RTX 3090 GPU with a batch size of \(40\).
We use the Adam optimizer with a cyclic learning rate schedule ranging from \(2 \times 10^{-5}\) to \(5 \times 10^{-4}\) over \(60\) epochs.
For the NOCS dataset, we use a \(3{:}1\) synthetic-to-real data ratio.

\subsection{Comparison with SOTA Methods}

\paragraph{Results on REAL275.}
Table~\ref{comparison} shows the quantitative results on the REAL275 dataset. The results demonstrate that our method achieves the best performance across all reported evaluation metrics. Specifically, under the evaluation settings of \(5^\circ 2\)~cm, \(5^\circ 5\)~cm, \(10^\circ 2\)~cm, and \(10^\circ 5\)~cm, TSM-Pose attains accuracies of 62.4\%, 68.5\%, 81.5\%, and 88.7\%, respectively. Compared with AG-Pose~\cite{lin2024instance}, which serves as the baseline, TSM-Pose yields consistent improvements of 8.6\%, 5.7\%, 8.5\%, and 4.7\% on the corresponding metrics. Furthermore, under the same evaluation protocol, TSM-Pose also outperforms the SOTA CleanPose method across all metrics. Figure~\ref{visual_comparision} presents qualitative comparisons, further illustrating that TSM-Pose produces more accurate and reliable pose estimates in complex scenarios.

\paragraph{Results on CAMERA25.}
Table~\ref{camera_comparison} reports the performance evaluation results on the CAMERA25 dataset. The experimental results indicate that TSM-Pose achieves the best performance across all evaluation metrics except IoU75 and the \(5^\circ 2\) cm criterion; even on these two metrics, its performance remains very close to that of the current SOTA methods. Specifically, TSM-Pose attains accuracy rates of 79.9\%, 84.6\%, 88.2\%,and 93.6\% under the \(5^\circ 2\)~cm, \(5^\circ 5\)~cm, \(10^\circ 2\)~cm, and \(10^\circ 5\)~cm metrics, respectively. Furthermore, the experimental results on the synthetic dataset further validate the effectiveness of the proposed TSM-Pose.

\paragraph{Results on HouseCat6D.}
As shown in Table~\ref{HouseCat6D_comparison}, our method consistently outperforms existing methods on HouseCat6D~\cite{jung2024housecat6d}. In particular, it improves the \(5^\circ\)2\,cm, \(10^\circ\)2\,cm, and \(10^\circ\)5\,cm scores by 5.4\%, 3.1\%, and 2.3\% over the best competing method, highlighting its strong generalization ability on COPE.

\begin{table}[t]
\centering
\small
\caption{Ablation study of individual modules. The best performance is highlighted in bold. TM denotes TwinMamba, TE denotes the Topology Extractor and SI denotes Semantic Injection.}
\label{ablation}
\resizebox{\linewidth}{!}{
\begin{tabular}{ccc rrrr}
\toprule
 TM & TE & SI 
& \(\mathbf{5^\circ\,2\,\mathrm{cm}}\)
& \(\mathbf{5^\circ\,5\,\mathrm{cm}}\)
& \(\mathbf{10^\circ\,2\,\mathrm{cm}}\)
& \(\mathbf{10^\circ\,5\,\mathrm{cm}}\) \\
\midrule
 \xmark & \xmark & \xmark & 57.0 & 64.6 & 75.1 & 84.7 \\
 \checkmark & \xmark & \xmark & 59.9 & 65.3 & 78.4 & 86.6 \\
 
 \checkmark & \checkmark & \xmark & 61.7 & 66.7 & 80.8 & 87.7 \\

 \checkmark & \checkmark & \checkmark & \textbf{62.4} & \textbf{68.5} & \textbf{81.5} & \textbf{88.7} \\
\bottomrule
\end{tabular}}
\end{table}

\begin{table}[t]
\centering
\small
\caption{Effect of serialization. Gray represents the default settings, and the best results are indicated in \textbf{bold}.}
\label{serialized_compare}

\renewcommand{\arraystretch}{1.2}
\resizebox{\columnwidth}{!}{
\begin{tabular}{lrrrr}
\toprule
Setting & \(\mathbf{5^\circ\,2\,\mathrm{cm}}\)
& \(\mathbf{5^\circ\,5\,\mathrm{cm}}\)
& \(\mathbf{10^\circ\,2\,\mathrm{cm}}\)
& \(\mathbf{10^\circ\,5\,\mathrm{cm}}\)\\
\midrule

Non-serialized & 60.8 & 66.7 & 79.8 & 87.5\\

Z-order      & 61.9 & 67.9 & 80.4 & 88.4\\

\rowcolor{gray!15}
Hilbert      & \textbf{62.4} & \textbf{68.5} & \textbf{81.5} & \textbf{88.7}\\

\bottomrule
\end{tabular}
}
\end{table}

\begin{table}[t]
\centering
\small
\caption{Effect of different mamba. Gray represents the default settings, and the best results are indicated in \textbf{bold}. CF denotes channel flipping, TF denotes token flipping, and RA denotes Random Channel Arrangement.}
\label{mamba_compare}

\renewcommand{\arraystretch}{1.1}
\resizebox{\columnwidth}{!}{
\begin{tabular}{lrrrr}
\toprule
Method & \(\mathbf{5^\circ\,2\,\mathrm{cm}}\)
& \(\mathbf{5^\circ\,5\,\mathrm{cm}}\)
& \(\mathbf{10^\circ\,2\,\mathrm{cm}}\)
& \(\mathbf{10^\circ\,5\,\mathrm{cm}}\)\\
\midrule
\rowcolor{gray!15}
TwinMamba      & \textbf{62.4} & \textbf{68.5} & \textbf{81.5} & \textbf{88.7}\\
BiMamba      & 61.2 & 67.3 & 77.4 & 85.6\\

Attention & 59.0 & 64.4 & 81.0 & 87.9\\
\midrule
\rowcolor{gray!15}
CF & \textbf{62.4} & \textbf{68.5} & \textbf{81.5} & \textbf{88.7}\\
TF & 60.9 & 66.6 & 79.5 & 87.0\\
RA & 58.2 & 65.3 & 77.4 & 86.5\\
\bottomrule
\end{tabular}
}
\end{table}

\begin{table}[t]
\centering
\small
\caption{Effect of different numbers of TwinMamba Blocks. Gray represents the default settings. The best results are indicated in \textbf{bold}.}
\label{TwinMambaNumber}

\renewcommand{\arraystretch}{1.0}
\resizebox{\columnwidth}{!}{
\begin{tabular}{rrrrr}
\toprule
Setting & \(\mathbf{5^\circ\,2\,\mathrm{cm}}\)
& \(\mathbf{5^\circ\,5\,\mathrm{cm}}\)
& \(\mathbf{10^\circ\,2\,\mathrm{cm}}\)
& \(\mathbf{10^\circ\,5\,\mathrm{cm}}\)\\
\midrule

2 & 60.4 & 66.3 & 80.0 & 87.7\\

4      & 61.1 & 67.3 & 80.2 & 88.2\\

\rowcolor{gray!15}
6      & \textbf{62.4} & \textbf{68.5} & \textbf{81.5} & \textbf{88.7}\\

8      & 60.7 & 67.0 & 78.0 & 85.8\\

10      & 60.5 & 65.9 & 79.7 & 86.5\\

\bottomrule
\end{tabular}
}
\end{table}

\subsection{Ablation Studies}
We conduct a systematic ablation study on a unified baseline model to analyze the contributions of three core components: TwinMamba, Topology Extractor, and Semantic Injection. We summarize the experimental results in Table~\ref{ablation}. We use AG-Pose as our baseline.

\paragraph{Effect of TwinMamba.}
As shown in Table~\ref{ablation}, the introduction of TwinMamba significantly improves the model's performance across all evaluation settings. At the \(10^\circ\)2cm setting, performance improves from 75.1\% to 78.4\%, and at the more stringent \(5^\circ\)2cm setting, performance improves from 57.0\% to 59.9\%, demonstrating that TwinMamba effectively captures long-range dependencies and globally models keypoints, significantly enhancing global feature representation. 
\paragraph{Effect of Topology Extractor.}
As shown in Figure~\ref{ablation}, the introduction of the Topology Extracto significantly improves model performance, increasing the accuracy at \(5^\circ\)2cm from 59.9\% to 61.7\%. By capturing global structural features that are robust to local geometric changes (as shown in Figure~\ref{feature}), the Topology Extracto supplements local geometric features and improves the consistency of intra-class features, thus improving the model's generalization ability.

\paragraph{Effect of Semantic Injection.}

As shown in Table~\ref{ablation}, with the introduction of Semantic Injection, the performance under the \(5^\circ\)2cm setting improves from 61.7\% to 62.4\%, indicating that incorporating semantic features effectively compensates for the semantic context lost during keypoint selection, thereby enhancing the semantic expressiveness of keypoints.

\paragraph{Effect of Serialization.}
In Table~\ref{serialized_compare}, we evaluated the impact of different serialization on the model. Non-serialized refers to the default point cloud order we used. Among the methods compared, Hilbert serialization achieved the best results.

\paragraph{Effect of Different Mamba.}
As shown in Table~\ref{mamba_compare}, BiMamba is instantiated from VisionMamba~\cite{vim}, while the Attention~\cite{vaswani2017attention} adopts standard self-attention with a comparable number of parameters.  TwinMamba consistently outperforms both BiMamba and the Attention baseline across all evaluation metrics. Notably, channel flipping yields significantly better performance than token flipping, indicating that preserving the spatial order of keypoints while constructing symmetric feature representations is more suitable for keypoint sequence modeling. In contrast, random channel arrangement (RA) performs worse, indicating that arbitrarily disrupting the channel order destroys all structural information in point features and is therefore detrimental to keypoint sequence modeling.

\paragraph{Effect of Varying Number of TwinMamba Blocks.}
As shown in Table~\ref{TwinMambaNumber}, increasing the number of blocks from 2 to 6 results in continuous improvement across all metrics, due to a deeper aggregation network that enhances the representation of global keypoints. However, further increasing the number of blocks from 6 to 10 may cause excessive global information fusion, resulting in homogenized features across different keypoints and reduced discriminative power. Overall, 6 blocks yield the best performance.

\paragraph{Note on Efficiency.}

Given that current topological feature extraction algorithms generally lack efficient GPU-parallel implementations, the introduction of topology-aware features limits the model throughput to a certain extent. Experimental results show that, during inference, TSM-Pose runs at 15 FPS, which is slightly lower than AG-Pose (23 FPS) and CleanPose (22 FPS). In terms of model size, TSM-Pose contains 38.35M parameters, which is higher than AG-Pose (33.47M) but still remains within the same order of magnitude overall. Although this introduces a certain inference overhead, TSM-Pose still maintains practical inference efficiency in offline analysis and batch-processing scenarios. More importantly, as shown in Table~\ref{comparison}, the proposed method achieves significant gains in robustness and generalization performance.

\section{Conclusion}

In this paper, we present TSM-Pose, a novel framework for COPE. Our approach introduces a Topology Extractor to explicitly capture global topological structures from point clouds, complemented by lightweight category tokens that inject global semantic priors into keypoints. Furthermore, the proposed TwinMamba module facilitates comprehensive global modeling, significantly bolstering the model's robustness against complex geometries and substantial intra-category variations. Extensive benchmarks demonstrate that TSM-Pose consistently surpasses state-of-the-art methods across multiple datasets. In the future, we aim to optimize the computational efficiency of topological feature generation through GPU parallelism and explore richer multimodal learning strategies to further enhance generalization.

\section*{Acknowledgements}
This research is supported by Zhejiang Provincial Natural Science Foundation of China under Grant No. LQN25F030009. We thank Wenhua Zhang and Liting Wang from the Cloud Network Technology Center at China Mobile Migu for providing computational resources through the Migu AI Training and Inference Service Platform.

\bibliographystyle{named}
\bibliography{ref}

\appendix
\clearpage

\begin{center}
\Large\textbf{Supplementary Material for Paper Titled \\``TSM-Pose: Topology-Aware Learning with Semantic Mamba for Category-Level Object Pose Estimation"}
\end{center}

\section{Instance-Adaptive Keypoint Detector }
The instance-adaptive keypoint detector (IAKD) follows the approach proposed in AG-Pose~\cite{lin2024instance}. It predicts a set of sparse yet discriminative keypoints. IAKD maintains a set of learnable keypoint queries $\mathbf{Q}=\{q_j\}_{j=1}^{N_k}$ shared across instances, and update them into instance-adaptive queries via an attention module:
\begin{equation}
\mathbf{Q}'=\mathrm{AttnLayer}(\mathbf{Q},\mathbf{F}),
\end{equation}
where $N_k$ is the number of keypoints.
Next, we compute the cosine similarity between each updated query $\mathbf{q}'_j$ and all point features $\mathbf{f}_i \in \mathcal{F}_f$, and apply a temperature-softmax to obtain a soft assignment heatmap:
\begin{equation}
h_{j,i}=
\frac{\exp\!\left(\cos(\mathbf{q}'_j,\mathbf{f}_i)/\tau\right)}
{\sum_{m=1}^{N}\exp\!\left(\cos(\mathbf{q}'_j,\mathbf{f}_m)/\tau\right)},
\quad \tau=0.1.
\end{equation}
Here, $h_{j, i}$ indicates the correspondence weight of point $i$ for keypoint $j$.
Finally, we compute the 3D location of each keypoint as the weighted expectation over point coordinates:
\begin{equation}
\mathbf{p}_{j}=\sum_{i=1}^{N} h_{j,i}\,\mathbf{x}_i,\quad \mathbf{x}_i \in \mathcal{P}_o,
\end{equation}
and \(\mathcal{P}_{kpt}=\{\mathbf{p}_j\}_{j=1}^{N_k}\).
Given the soft assignment heatmap $h_{j,i}$, we let $\mathbf{f}_i$ denote the fused feature of the $i$-th point, then the feature of the $j$-th keypoint is defined as:
\begin{equation}
\mathbf{f}^{kpt}_j=\sum_{i=1}^{N} h_{j,i}\,\mathbf{f}_i ,
\end{equation} 
Then, we inject a positional encoding derived from the predicted keypoint coordinates:
\begin{equation}
\tilde{\mathbf{f}}^{kpt}_j = \mathbf{f}^{kpt}_j + \phi(\mathbf{p}_j),
\end{equation}
where $\phi(\cdot)$ is a learnable MLP that maps the 3D keypoint location $\mathbf{p}_j \in \mathbb{R}^3$ to the feature dimension, and \(\mathcal{F}_{kpt}=\{\tilde{\mathbf{f}}^{kpt}_j\}_{j=1}^{N_k}\)

To ensure geometrically meaningful keypoints, we further encourage the predicted keypoints to lie close to the object surface.
Given the object surface point cloud \(\mathcal{P}_{o}=\{\mathbf{x}_i\}_{i=1}^{N}\), we adopt a one-sided, object-aware Chamfer distance:

\begin{equation}
\mathcal{L}_{surf}
= \frac{1}{N_k}
\sum_{\mathbf{p}_i\in\mathcal{P}_{kpt}}
\min_{\mathbf{x}_j\in\mathcal{P}_{o}}
\|\mathbf{p}_i - \mathbf{x}_j\|_2.
\end{equation}

This formulation effectively suppresses outlier keypoints while maintaining stable optimization. Moreover, to prevent keypoint collapse and encourage spatial diversity, we introduce a margin-based diversity loss:
\begin{equation}
\mathcal{L}_{div}
= \sum_{i=1}^{N_k}\sum_{j=1, j\neq i}^{N_k}
\max\left(0, th - \|\mathbf{p}_i - \mathbf{p}_j\|_2\right),
\end{equation}
where \(th = 0.01\) is a predefined distance threshold measured in meters.
This loss penalizes pairs of keypoints that are closer than \(th\), encouraging their dispersion over the object surface.

This differentiable selection process enables instance-adaptive keypoint localization, providing stable structural cues for subsequent pose regression.

\section{Loss Function Details}


There are additional loss functions used to constrain keypoint and pose estimation.
Let \(\mathcal{P}^{nocs}_{kpt}=\{\mathbf{\hat{u}}_i\}_{i=1}^{N_k}\) denote their corresponding predicted coordinates in the NOCS.
Using the ground-truth pose parameters, we project each keypoint into the canonical space as:
\begin{equation}
\mathbf{u}^{gt}_i
= \frac{1}{\|\mathbf{s}_{gt}\|_2} R_{gt}(\mathbf{p}_i - \mathbf{t}_{gt}),
\end{equation}
which yields the ground-truth NOCS keypoints \(\mathcal{P}^{gt}_{kpt}=\{\mathbf{u}^{gt}_i\}\).
Consistency between predicted and ground-truth NOCS coordinates is enforced using a Smooth L1 loss:
\begin{equation}
\mathcal{L}_{nocs}
= \frac{1}{N_k}\sum_{i=1}^{N_k}
\mathrm{SmoothL1}(\mathbf{u}^{gt}_i, \mathbf{\hat{u}}_i).
\end{equation}
This auxiliary supervision stabilizes category-level pose learning across intra-class shape variations. 

\begin{table}[t]
\renewcommand{\arraystretch}{1.25}
\caption{Overall and category-wise evaluation on REAL275.
We highlight the best results in \textbf{bold}. '*' denotes CATRE IoU metrics.}
\label{real}
\setlength{\tabcolsep}{4pt}
\resizebox{\linewidth}{!}{%
\begin{tabular}{l l rrrrr}
\toprule
\textbf{Object} & \textbf{Method}
& IoU$_{75}$* 
& \(\mathbf{5^\circ\,2\,\mathrm{cm}}\)
& \(\mathbf{5^\circ\,5\,\mathrm{cm}}\)
& \(\mathbf{10^\circ\,2\,\mathrm{cm}}\)
& \(\mathbf{10^\circ\,5\,\mathrm{cm}}\) \\
\midrule

\multirow{3}{*}{Bottle}
& AG-Pose    & 36.8 & 72.8 & 80.0 & 79.1 & 87.8 \\
& CleanPose  &  37.1  & 75.7   & 81.7   & 79.9   & 87.8   \\
& Ours       & \textbf{38.3} & \textbf{78.3} & \textbf{83.6} & \textbf{82.0} & \textbf{87.9} \\
\midrule
\multirow{3}{*}{Bowl}
& AG-Pose    & 94.0 & \textbf{93.8} & \textbf{98.9} & 94.7 & 99.9 \\
& CleanPose  &   93.8  & 93.3  & 98.2   & 95.0  & 99.9   \\
& Ours       & \textbf{94.0} & 93.5 & 98.1 & \textbf{95.5} & \textbf{100.0} \\
\midrule

\multirow{3}{*}{Camera}
& AG-Pose    & 36.7 & 1.4 & 1.7 & 28.7 & 34.0 \\
& CleanPose  & \textbf{39.9}   & 2.8  & 3.2  & 33.9  & 40.5  \\
& Ours       & 36.2 & \textbf{3.1} & \textbf{3.3} & \textbf{49.0} & \textbf{53.3} \\
\midrule

\multirow{3}{*}{Can}
& AG-Pose    & 38.6& 80.7 & 83.7 & \textbf{97.0} & \textbf{99.8} \\
& CleanPose  & 43.2  & \textbf{84.2}   & \textbf{85.9}   & 96.9  & 98.6   \\
& Ours       & \textbf{44.9} & 82.3 & 85.6 & 96.6 & 98.7 \\
\midrule

\multirow{3}{*}{Laptop}
& AG-Pose    & \textbf{78.3} & 59.8 & 89.7 & 63.1 & 97.6 \\
& CleanPose  & 76.1   & 69.2   & 90.9   & 71.9   & \textbf{98.5}  \\
& Ours       & 78.1 & \textbf{70.5} & \textbf{93.9} & \textbf{72.6} & 98.3 \\
\midrule

\multirow{3}{*}{Mug}
& AG-Pose    & 82.9 & 34.6 & 34.8 & 89.5 & 90.0 \\
& CleanPose  & 86.1   &45.2  & 45.6  & 91.9  & 91.9   \\
& Ours       & \textbf{86.8} & \textbf{46.1} & \textbf{46.5} & \textbf{93.3} & \textbf{93.9} \\
\midrule

\multirow{3}{*}{Average}
& AG-Pose    & 61.3 & 57.0 & 64.6 & 75.1 & 84.7 \\
& CleanPose  & 62.7 & 61.7 & 67.6 & 78.3 & 86.3 \\
& Ours       & \textbf{63.1} & \textbf{62.4} & \textbf{68.5} & \textbf{81.5} & \textbf{88.7} \\

\bottomrule
\end{tabular}}
\end{table}

\section{Topological Feature Analysis}
\begin{figure*}
\centering
\setlength{\tabcolsep}{1pt}
\renewcommand{\arraystretch}{0.5}
\scriptsize 
\begin{tabular}{ccc}
    \includegraphics[width=0.32\textwidth]{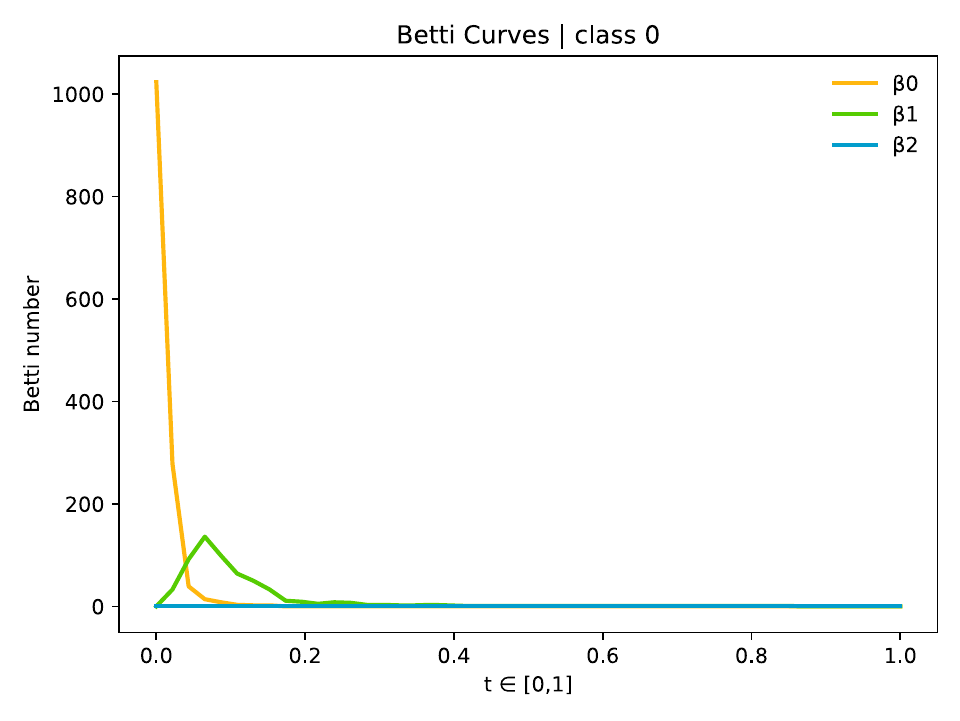}&
    \includegraphics[width=0.32\textwidth]{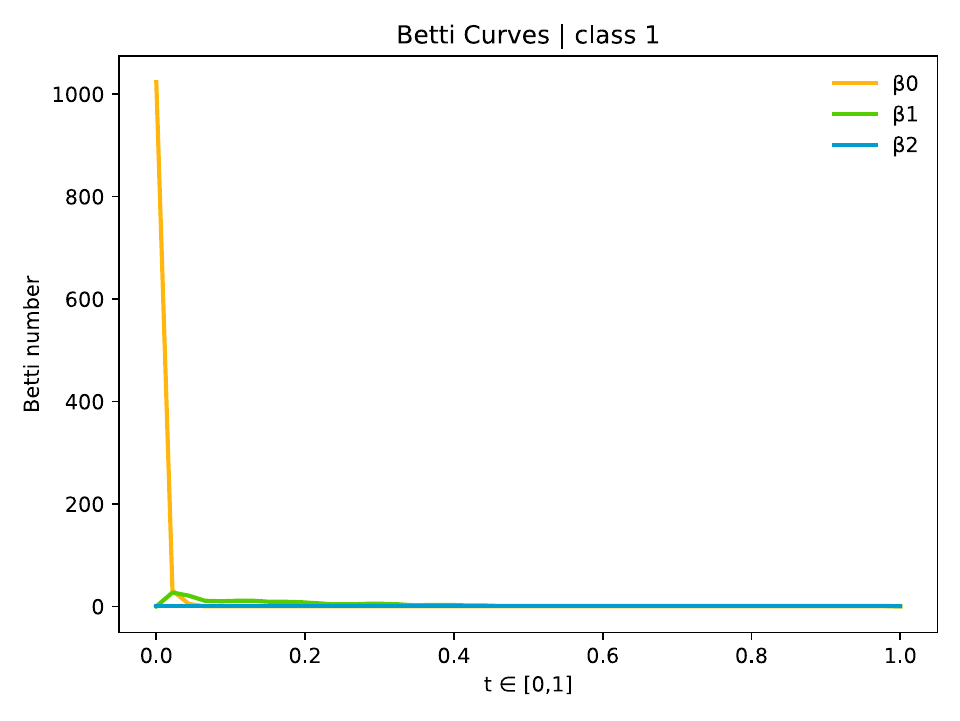}&
    \includegraphics[width=0.32\textwidth]{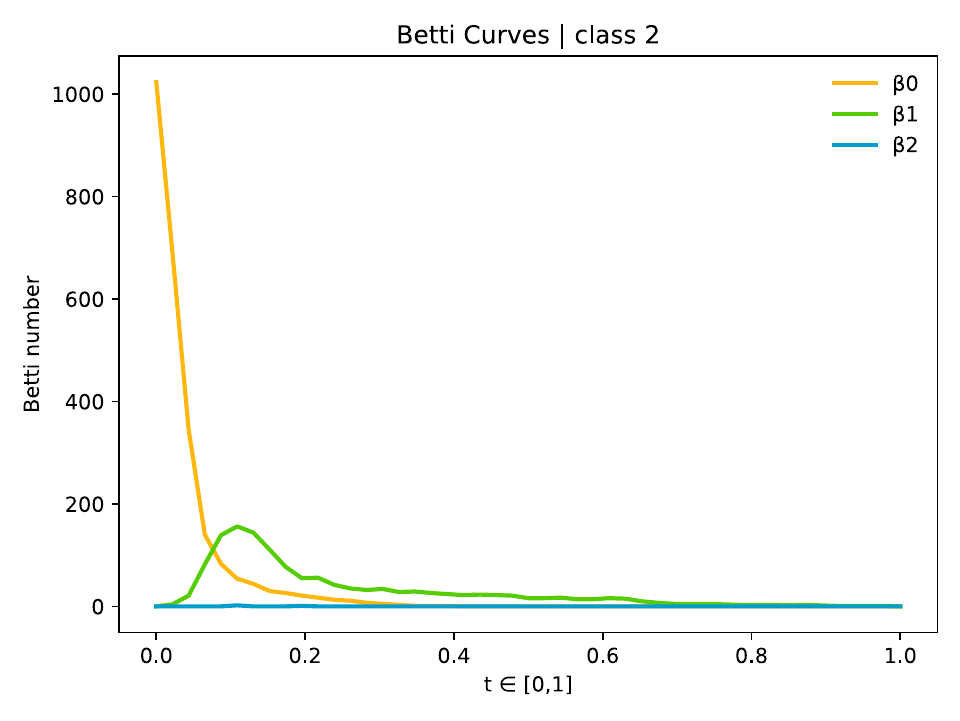}\\
    \includegraphics[width=0.32\textwidth]{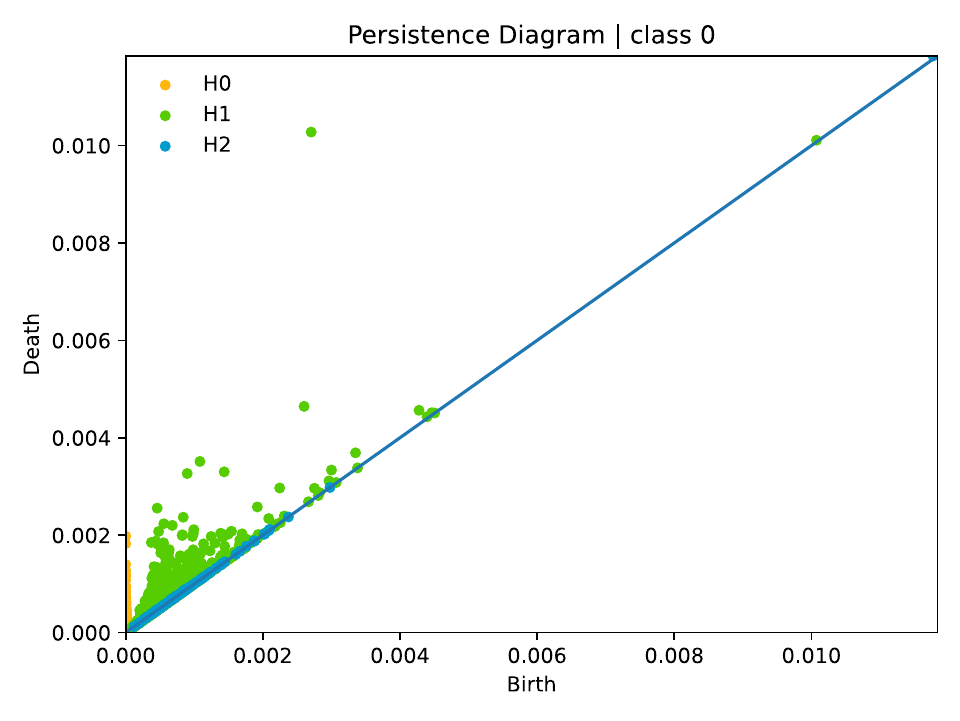}&
    \includegraphics[width=0.32\textwidth]{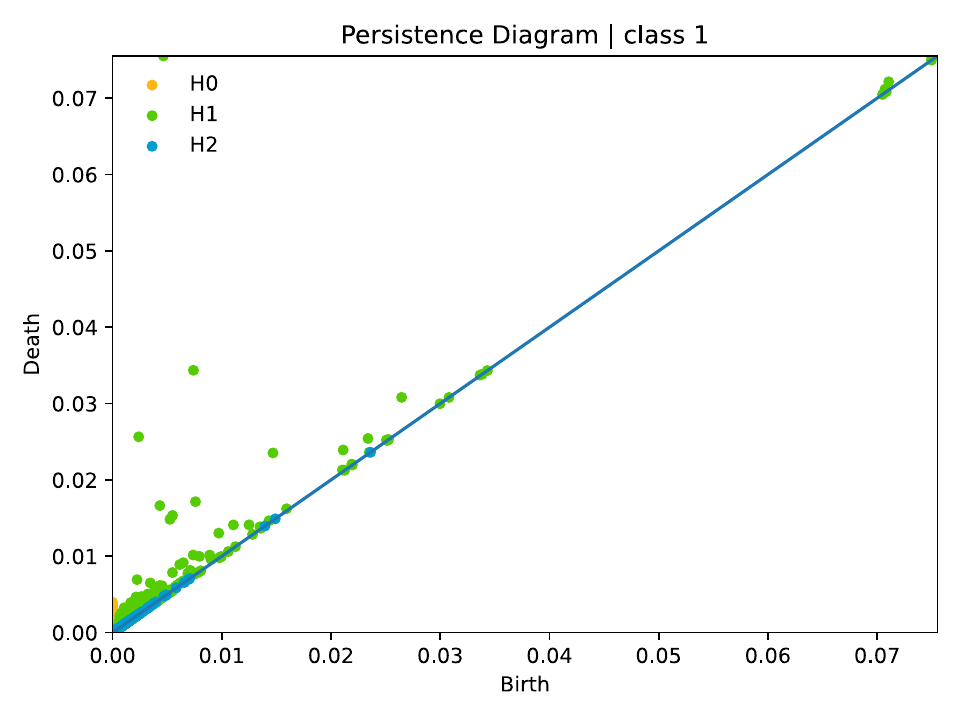}&
    \includegraphics[width=0.32\textwidth]{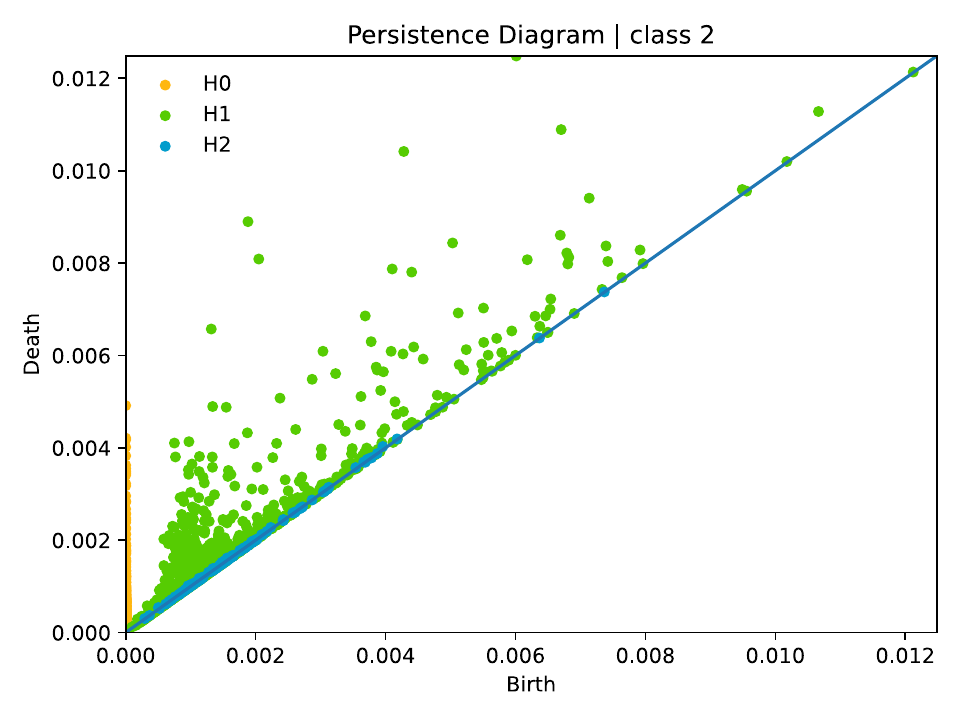}\\
     \textbf{(a) bottle}& \textbf{(b) bowl} & \textbf{(c) camera} \\
    \includegraphics[width=0.32\textwidth]{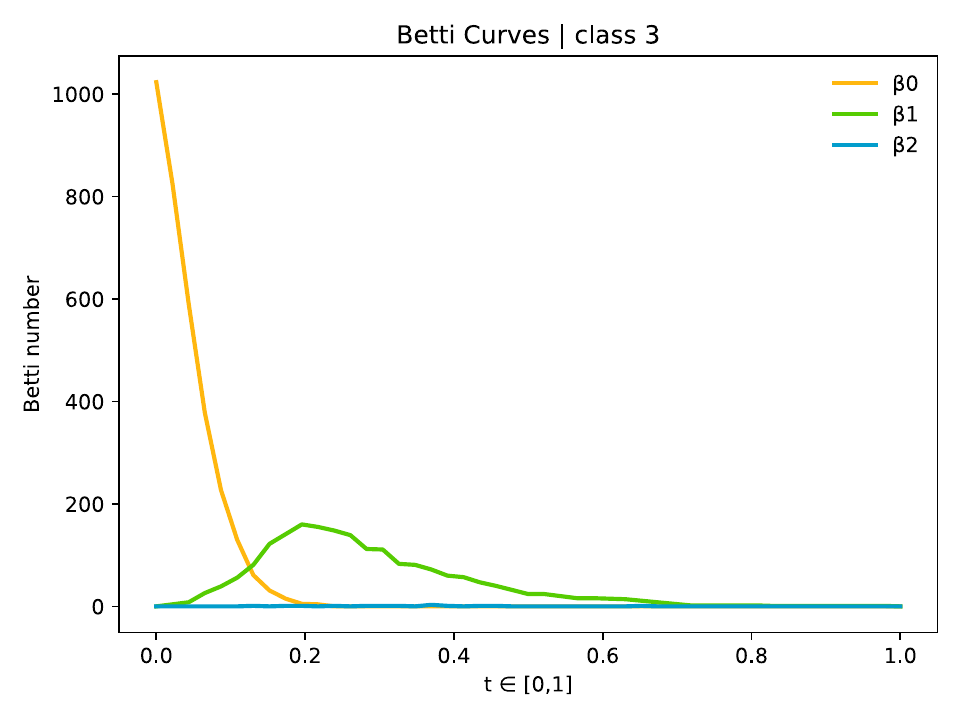}&
    \includegraphics[width=0.32\textwidth]{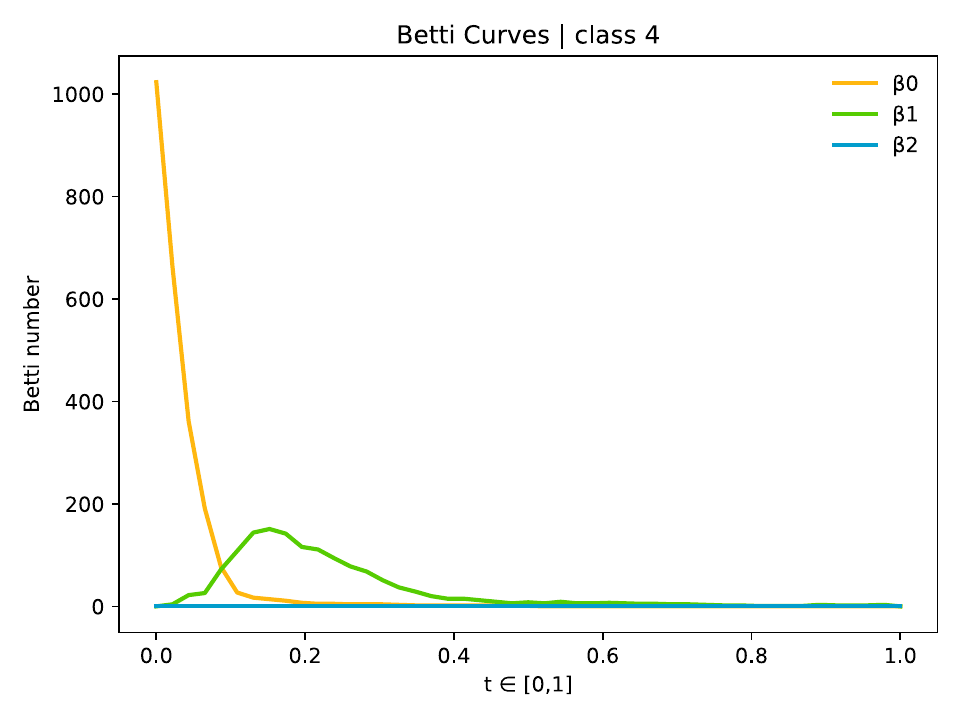}&
    \includegraphics[width=0.32\textwidth]{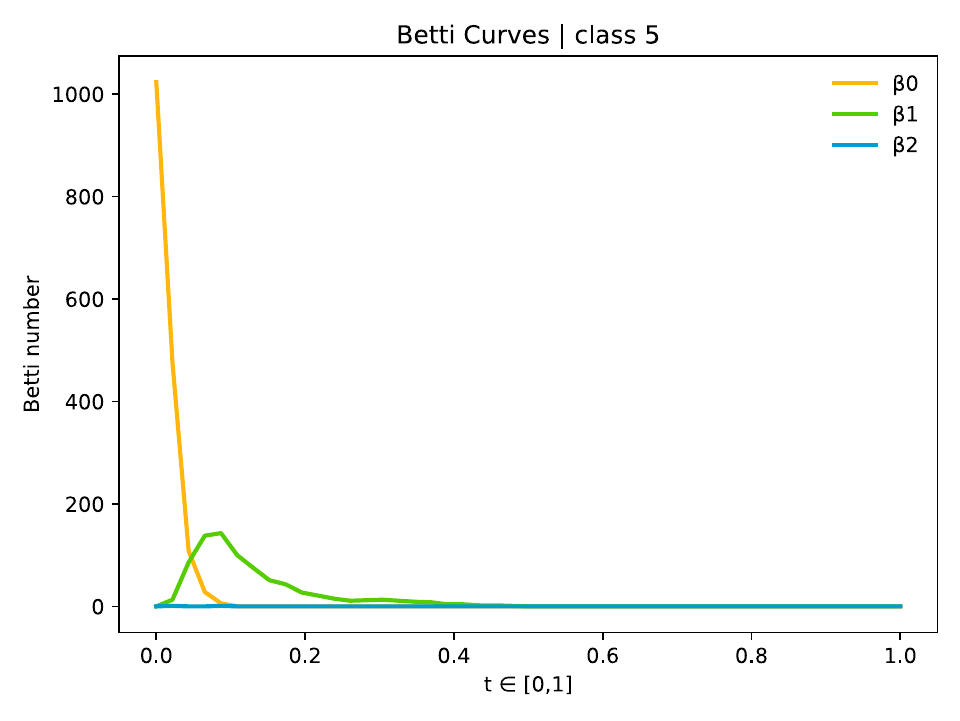}
    \\
    \includegraphics[width=0.32\textwidth]{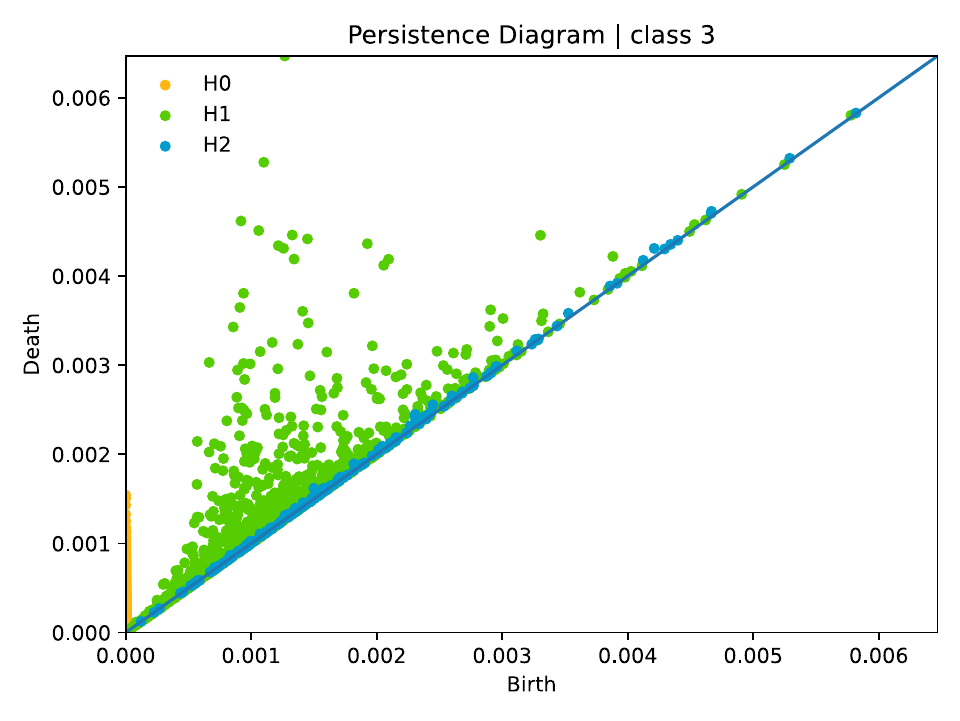}&
    \includegraphics[width=0.32\textwidth]{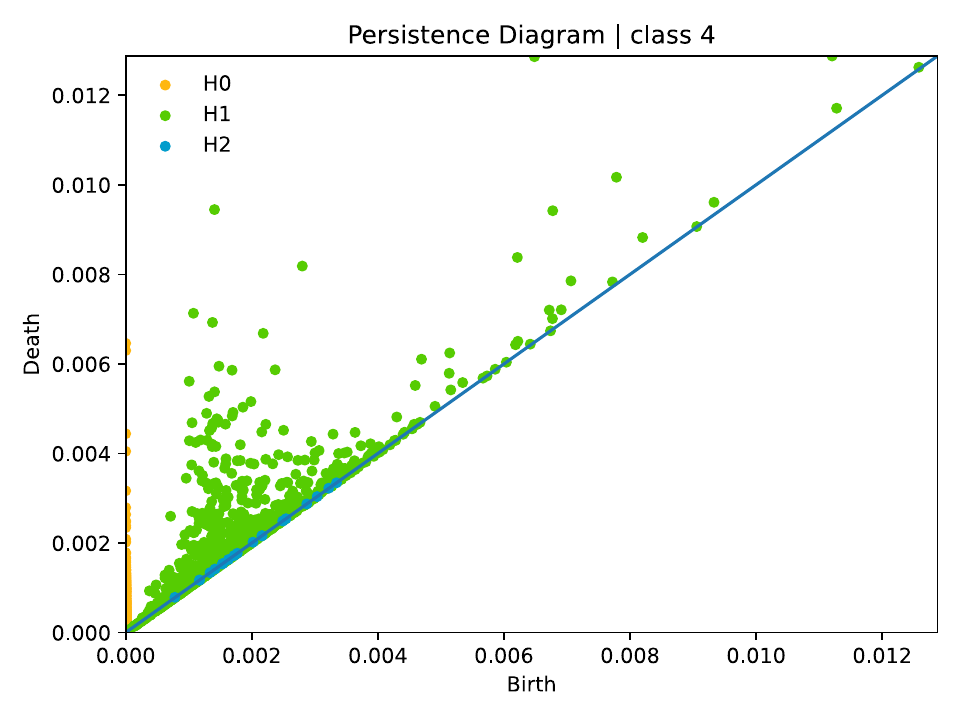}&
    \includegraphics[width=0.32\textwidth]{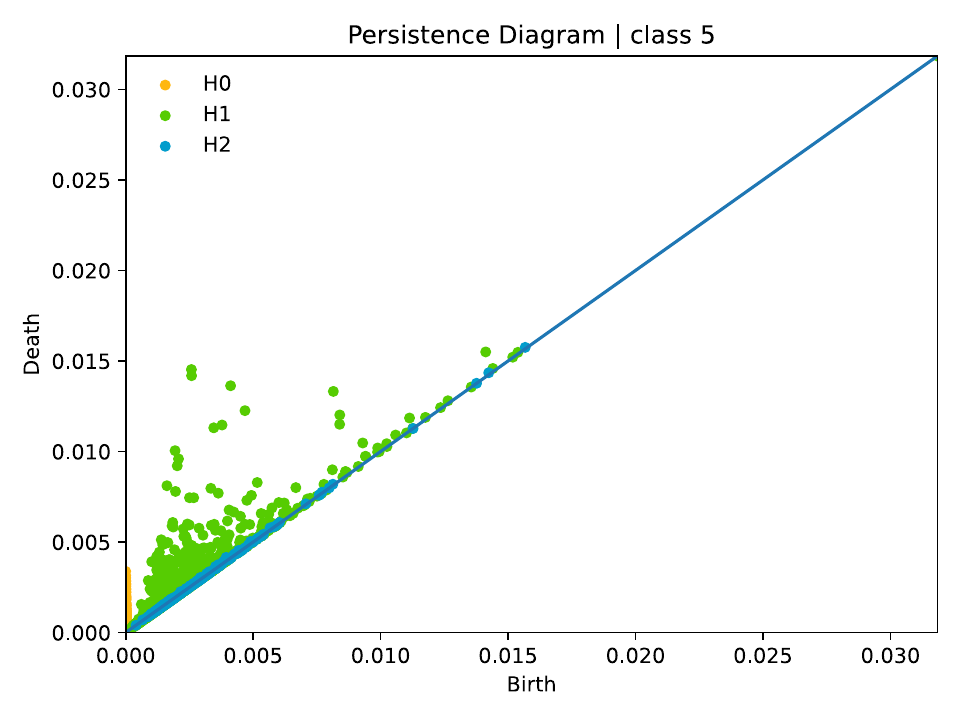}\\
   \textbf{(d) can}& \textbf{(e) laptop} & \textbf{(f) mug}  \\
\end{tabular}
\caption{Topological representations of six object categories. For each category, the Betti curves (top) and the corresponding persistence diagrams (bottom) are shown.}
\label{topovis}
\end{figure*}

We visualize the persistent homology representations of six object categories, including Betti curves and corresponding persistence graphs, in Figure~\ref{topovis}.

Betti curves characterize the trend in the number of topological features as the scale parameter varies at different homology dimensions. The horizontal axis represents the evolution of the filtering scale from local to global, and the vertical axis represents the number of Betti curves for the corresponding homology dimension at that scale. Betti curves provide a visual observation of differences in topological complexity and evolution across categories.

The persistence diagram characterizes the topological structure from the perspective of feature lifetimes. Each point in the persistence diagram corresponds to a topological feature, with the horizontal and vertical axes representing the scales at which the feature appears and disappears, respectively. Points farther from the diagonal represent longer-lived and more stable features, while points closer to the diagonal typically correspond to transient, noisy structures.

In our experimental setup, the input point clouds primarily originate from single- or limited-view surface scans. Consequently, their geometry typically covers only the outer surfaces of objects, making it challenging to form closed volumes or complete cavity structures. Therefore, the topological features corresponding to the H2 dimension appear infrequently in most object categories, and their lifetimes are usually short, mainly distributed near the diagonal of the persistent graph. These features are sensitive to sampling density and noise, and provide relatively limited information for class-level structural differentiation.

In contrast, the H0 and H1 dimensions can more stably characterize the main topological features of objects in terms of connectivity and toroidal structures, and exhibit better repeatability and consistency across different instances. Therefore, this paper primarily focuses on topological representations of the H0 and H1 dimensions to achieve a good balance among expressive power, stability, and computational efficiency.

\section{Extra Experimental Results on REAL275}
As shown in Table~\ref{real}, we conduct detailed quantitative evaluations across different categories using the REAL275 dataset. We compare our method with the baseline AG-Pose and the current state-of-the-art CleanPose. Overall, our method achieves state-of-the-art performance on average, outperforming the comparison methods on five evaluation metrics: IoU$_{75}$, \(5^\circ\,2\,\mathrm{cm}\), \(5^\circ\,5\,\mathrm{cm}\), \(10^\circ\,2\,\mathrm{cm}\), and \(10^\circ\,5\,\mathrm{cm}\), validating its overall advantage in COPE.

From a category-level perspective, our method shows consistent improvement on objects with regular geometric structures such as Bottles, Bowls, and Mugs, with a particularly significant advantage under more stringent pose evaluation criteria. For the Camera category, although the IoU\(_{75}\) is slightly lower than the state-of-the-art method, significant improvements of 49.0\% and 53.3\% are achieved under more relaxed pose constraints (\(10^\circ\,2\,\mathrm{cm}\) and \(10^\circ\,5\,\mathrm{cm}\)). Furthermore, in categories with complex component structures, such as Laptop and Can, our method achieves state-of-the-art or near-optimal results on most metrics, further validating its effectiveness.

\end{document}